\documentclass[11pt]{article}

\usepackage[preprint]{acl}

\usepackage{times}
\usepackage{booktabs}
\usepackage{latexsym}
\usepackage{subfigure}
\usepackage{enumitem}
\usepackage{float}
\usepackage{amsmath}
\usepackage{tabularx}
\usepackage{float}
\usepackage{multicol, multirow}
\usepackage{tcolorbox}
\usepackage{enumitem}
\usepackage{textcomp}
\usepackage{xspace}
\usepackage{array}
\usepackage{makecell}
\usepackage{arydshln}
\usepackage{bookmark}
\usepackage{adjustbox}
\usepackage{xurl}
\usepackage{colortbl}  
\usepackage{tikz}      
\usepackage{soul}
\usepackage{afterpage}
\usepackage{fancyhdr}
\usepackage{bm}
\tcbuselibrary{skins}

\usepackage[T1]{fontenc}

\usepackage[utf8]{inputenc}

\usepackage{microtype}

\usepackage{inconsolata}

\usepackage{graphicx}

\newcommand{\eg}{{\textit{e.g.}}}

\newcommand{\rom}[1]{\lowercase\expandafter{\romannumeral #1\relax}}

\newcommand\red[1]{\textcolor{red}{#1}}

\definecolor{MySoftGreen}{RGB}{34,139,34}
\definecolor{MyPurple}{RGB}{104, 52, 154}
\definecolor{baselineColor}{RGB}{30, 80, 160} 
\definecolor{ronaColor}{RGB}{180, 50, 50}     

\newcolumntype{Y}{>{\centering\arraybackslash}X}
\newcolumntype{L}{>{\raggedright\arraybackslash}X}

\definecolor{darkblue}{rgb}{0, 0, 0.5}
\hypersetup{colorlinks=true, citecolor=darkblue, linkcolor=darkblue, urlcolor=darkblue}

\title{Generative Active Testing: Efficient LLM Evaluation via Proxy Task Adaptation}

\author{Aashish Anantha Ramakrishnan\textsuperscript{1}\thanks{\ Majority of work done during an internship at Optum AI.}~~~~
Ardavan Saeedi\textsuperscript{2}~~~~
Hamid Reza Hassanzadeh\textsuperscript{2}\\
\textbf{Fazlolah Mohaghegh\textsuperscript{2}~~~~
Dongwon Lee\textsuperscript{1}}\\[0.5em]
  \textsuperscript{1}The Pennsylvania State University~~~~
  \textsuperscript{2}Optum AI\\
\textsuperscript{1}\texttt{\{aza6352,dul13\}@psu.edu}~~~\\
\textsuperscript{2}\texttt{\{ardavan.saeedi,hamid.hassanzadeh,ehsan.mohaghegh\}@optum.com}\\
}

\begin{document}
\maketitle

\begin{abstract}
With the widespread adoption of pre-trained Large Language Models (LLM), there exists a high demand for task-specific test sets to benchmark their performance in domains such as healthcare and biomedicine. However, the cost of labeling test samples while developing new benchmarks poses a significant challenge, especially when expert annotators are required. Existing frameworks for active sample selection offer limited support for generative Question Answering tasks, where option dynamics can affect model decision boundaries. In this paper, we present Generative Active Testing (GAT), an uncertainty-aware acquisition framework leveraging LLMs as surrogates for informing the sample selection process. Using a novel Statement Adaptation Module, we modify generative tasks into a pseudo-classification format, enabling the capture of sample-level uncertainties across unlabeled candidates. Our zero-shot acquisition functions reduce estimation error by $\sim$40\% compared to traditional sampling baselines, offering a scalable solution for cost-effective model benchmarking.
\end{abstract}

\begin{figure*}
      \centering
      \includegraphics[width=0.7\linewidth]{"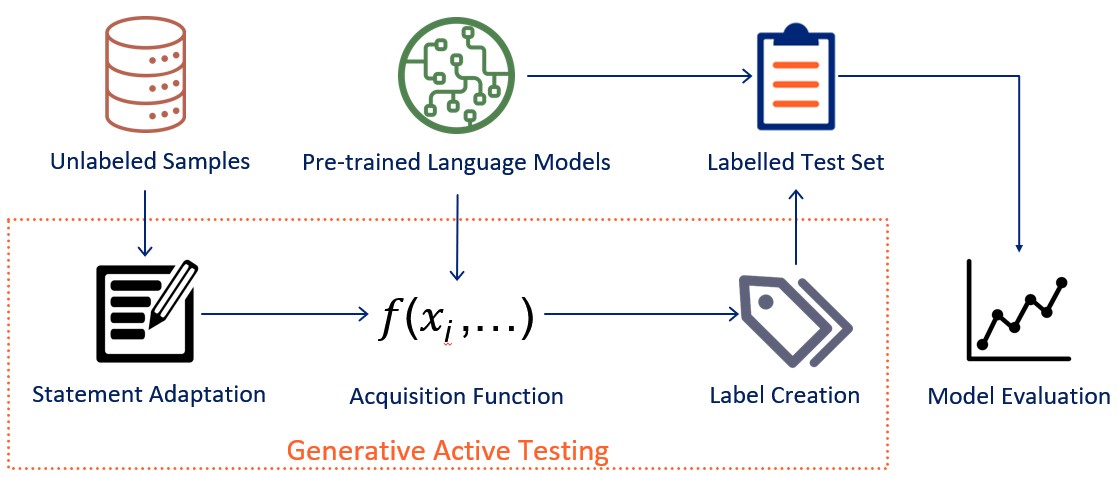"}
      \caption{An \textbf{overview} of the components of the Generative Active Testing Framework.}
      \label{fig:overview}
\end{figure*}

\section{Introduction}

The rapid proliferation of Large Language Models (LLMs) has necessitated robust benchmarks to validate their capabilities across diverse tasks. While comprehensive benchmarks like MMLU \cite{Hendrycks2020-bc} and HellaSwag \cite{Zellers2019-xv} provide valuable insights into factual and commonsense reasoning, their increasing scale creates a significant bottleneck: evaluation cost. This challenge is particularly acute in specialized high-stakes domains such as biomedicine and health informatics. In tasks such as Medical Prior Authorization, determining coverage eligibility often relies on complex clinical questionnaires \cite{Vatsal2024-he, Pandey2024-ja} structured as a series of context-based Multiple-Choice Question Answers (MCQA). Validating models on these workflows requires gold-standard labels from expert annotators ({\eg} clinicians), whose time is limited and prohibitively expensive compared to compute resources \cite{Kossen2022-dr}. Consequently, annotated data is strictly prioritized for training, leaving a scarcity of high-quality data for reliable evaluation.

To address this, we turn to \textit{Active Testing} \cite{Sawade2010-cr, Kossen2021-xc}, an Active Learning-inspired paradigm for acquiring the most informative test samples to estimate model performance. A key component of Active Testing is the use of "surrogate" models to guide \textit{acquisition functions} by estimating uncertainty. However, standard MCQA tasks lack a fixed label space, confining uncertainty estimates to the local context of each question.  This fragmentation prevents meaningful comparison of confidence scores between different samples, making it challenging to gauge the model's grasp of the broader data distribution. Option dynamics also play a crucial role where confusing candidates can serve as distractors, potentially increasing model uncertainty. We investigate two core research questions:

\vspace{0.2cm}
\noindent \textbf{RQ1:} \textit{Can we transform dynamic MCQA tasks into a fixed proxy that makes uncertainty comparable across different questions, enabling a reliable estimate of the model's grasp of the global data distribution?}

\vspace{0.1cm}
\noindent \textbf{RQ2:} \textit{How can we design LLM-based acquisition functions that leverage model uncertainty to select informative samples more effectively than traditional baselines?}
\vspace{0.2cm}

We propose a \textbf{Generative Active Testing} framework that transforms dynamic MCQA tasks into a fixed \textit{Statement Verification} task. This adaptation converts generative outputs into a pseudo-classification format (True/False), satisfying the requirements for unbiased risk estimation using different \textit{oracle strategies} for option selection. Finally, we propose LLM-based acquisition functions that leverage surrogate model distributions to estimate sample informativeness through entropy regularization. These "prior" distributions represents a reference belief about possible outcomes before incorporating the specific confidence estimates of the main model being evaluated. Crucially, our approach operates in a zero-shot manner, requiring no task-specific fine-tuning. This eliminates the high computational overhead typically associated with training custom uncertainty estimators, making our framework highly scalable and easy to integrate into existing pipelines compared to baselines like Random Sampling. This efficiency gain offers a viable pathway for reducing expert annotation costs in healthcare applications without compromising benchmark reliability or incurring additional training debt.

\section{Related Work}
The choice of acquisition functions for sample selection is critical to the success of model-efficient \cite{Zhao2024-da, Sener2017-je} and data-efficient \cite{Su2024-bl, Yu2023-ag, Ochiai2023-ib} evaluation approaches. However, existing active acquisition functions have only been extensively evaluated using machine learning models trained on predictive tasks such as Regression or Classification. Traditional sampling methods such as Random Sampling may provide an unbiased selection of samples, but suffer from high variance between selection runs, leading to higher estimation errors on average \cite{Kossen2021-xc}. Clustering-based sampling methods may provide a more representative selection of samples, but may introduce bias in the risk estimates due to over-reliance on data diversity features \cite{Ashury-Tahan2024-up}. The computational cost of extracting cluster labels also limits its usefulness for large-scale sample selection. The use of LLMs for sample selection has been studied for classification-style tasks \cite{Berrada2025-ds}. However extending them to true generative-style tasks such as QA have been limited, with the inherent task complexity leading to challenges in understanding their representations effectively \cite{Huang2026-we}. Thus, there exists \textit{the need for novel acquisition functions that adapt well to generative tasks}.

\section{Methodology}
\subsection{Task Definition}
Consider a model $\mathcal{F}: \mathcal{X} \rightarrow \mathcal{Y}$, where $x \in \mathcal{X}$ is a sample we wish to evaluate on our task. Active Testing is a framework that aims to select a subset of samples ($\mathcal{D}_{sub}$) from a large pool of candidates ($\mathcal{D}_{test}$) to evaluate the performance of a selected model. In a discriminative task, $\mathcal{D}_{test}$ contains $\{(x_1, y_1), (x_2, y_2), ..., (x_N, y_N)\}$, where $x_i$ and $y_i$ correspond to each data point and its output label respectively where $i\le{N} $. The Empirical Risk $\hat{R}$, which is the average loss $\mathcal{L}$ across the entire candidate set, is used to evaluate the model's estimation error on $\mathcal{D}_{test}$. A key constraint that drives the selection of samples is the sampling budget $M$, which limits the number of samples that can be acquired for evaluation \cite{Sawade2010-cr}. This sampling budget is representative of real-world constraints such as annotation costs or time constraints for model evaluation. When we consider a subset $\mathcal{D}_{sub}$ selected from $\mathcal{D}_{test}$, the Empirical Risk of the subset $\hat{R}_{sub}$ is defined as:

\begin{equation}
      \hat{R}_{sub} = \frac{1}{M} \sum_{i=1}^{M} \mathcal{L}(y_i, \mathcal{F}(x_i))
\end{equation}

The goal of Active Testing is to select samples that are most informative for model evaluation at every sampling budget, while still maintaining the representativeness of the data distribution. This approach is different from adaptive testing \cite{Polo2024-fu, Zhuang2023-ab}, where the objective is to reduce the number of samples required to assess a particular task when having access to labeled samples. Here, $\hat{R}_{sub}$ should be an unbiased estimate of the model's estimation error when using only the subset $\mathcal{D}_{sub}$ for evaluation with $\hat{R}_{sub} \rightarrow \hat{R}$ as $M \rightarrow N$. Existing unbiased estimators such as the Levelled Unbiased Risk Estimator ($\hat{R}_{LURE}$) \cite{Farquhar2021-rs} and Active Surrogate Estimators \cite{Kossen2022-dr} have been shown to provide good estimates of the model's estimation error when using only a subset of the original data. However, the format of generative tasks such as QA may not be directly extendable to these estimators. 

\begin{figure*}
      \centering
      \includegraphics[width=0.7\linewidth]{"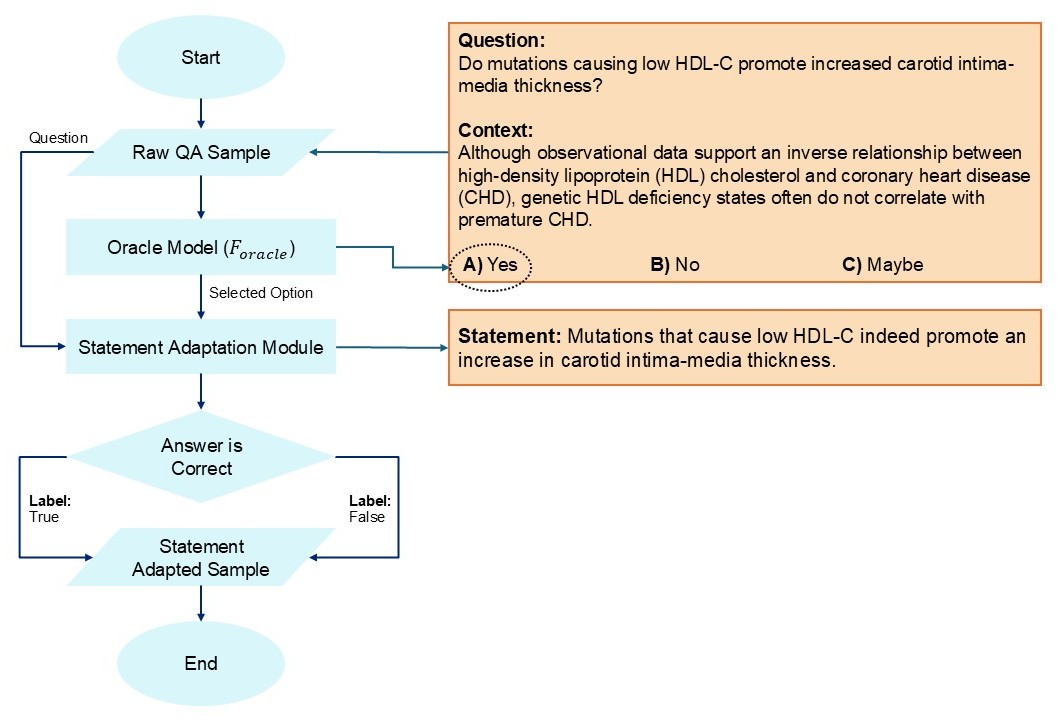"}
      \caption{\textbf{Step-wise flow diagram} detailing tasks performed by the Statement Adaptation Module.}
      \label{fig:adaptation}
\end{figure*}

\subsection{Generative Active Testing: An Overview}
Our proposed GAT framework consists of 2 key components: \textbf{Statement Adaptation} and \textbf{Acquisition Functions}. The Statement Adaptation module enables the conversion of dynamic Natural Language Generation tasks, such as Question Answering, into a fixed-schema classification setting. Through this Natural Language Verification (NLV) task, we obtain a constrained set of binary classes (True/False) against which the model's uncertainty can be standardized. Using this fixed output space, we devise LLM-based acquisition functions for efficient sample selection. These acquisition functions leverage the \textit{entropy distributions} of the surrogate models on the unlabeled test set to inform the sampling process. Finally, we compare the performance of LLMs across these different acquisition functions to evaluate their suitability in minimizing the estimation error of the global risk.

\paragraph*{Statement Adaptation}
To deploy unbiased risk estimators for Active Testing, the target task must operate over a fixed, context-independent label space ({\eg} discrete classes). Standard Question Answering (QA) tasks violate this requirement, as output options vary dynamically per input \cite{Jin2019-au,Clark2019-ns}. To resolve this, we propose a \textit{Statement Adaptation} module (Figure \ref{fig:adaptation}) that transforms the QA task into a binary Natural Language Verification (NLV) task with fixed labels \{True, False\} \cite{Sanyal2024-nj, Chen2021-by}.

Crucially, our evaluation targets the high-confidence regime ($\ge 0.7$), where the model's native uncertainty often fails to distinguish between correct predictions and "silent failures" (hallucinations) \cite{Ishii2025-gl}. These errors are particularly dangerous in biomedical settings, as they bypass standard low-confidence review filters. To recover a usable error signal in this regime, we construct proxy statements using an oracle model $\mathcal{F}_{oracle}$ (Figure \ref{fig:adaptation}) and investigate three probing strategies: \textit{MostConf} (targeting the primary prediction), \textit{LeastConf} (lowest-probability option), and \textit{RunnerUp} (second most probable option).  To assess the efficacy of these strategies, we calculate the Conditional Area Under the Receiver Operating Characteristic curve (Cond. AUROC). Unlike standard metrics, this specifically measures the proxy task's discriminative power within the high-confidence subset quantifying its ability to "unmask" hallucinations by separating them from robust knowledge.

\begin{table}[h]
\centering
\footnotesize
\setlength{\tabcolsep}{0pt} 
\renewcommand{\tabularxcolumn}[1]{m{#1}}
\begin{tabularx}{\linewidth}{@{\extracolsep{\fill}} l Y Y Y @{}}
\toprule
& \multicolumn{3}{c}{\textbf{$\Delta$ Conditional AU-ROC (Conf $\ge0.7$)}} \\
\cmidrule(lr){2-4}
\textbf{Strategy} & \textbf{PubMedQA} & \textbf{MedQA} & \textbf{AI2\_ARC} \\
\midrule
\midrule
MostConf & +3.9\% & \textcolor{red}{-6.1\%} & \textcolor{red}{-49.3\%} \\
LeastConf & \textcolor{red}{-49.9\%} & +25.6\% & +11.0\% \\
\textbf{RunnerUp} & \textbf{+2.1\%} & \textbf{+4.9\%} & \textbf{+3.3\%} \\
\midrule
\midrule
\textit{Subset Size ($N$)} & 352 (70\%) & 394 (31\%) & 363 (59\%) \\
\bottomrule
\end{tabularx}
\caption{\textbf{Ablation of Option Selection Strategies.} Values represent the change ($\Delta$) in Conditional AUROC (Conf $\ge0.7$), with the \textit{RunnerUp} strategy showing consistent, robust improvements across all datasets.}
\label{tab:strategy_ablation_vertical}
\end{table}

As shown in Table \ref{tab:strategy_ablation_vertical}, strategies like \textit{MostConf} and \textit{LeastConf} exhibit high variance due to dataset-specific option dynamics where the presence of strong distractors or ambiguous "Maybe" options can skew the proxy task's difficulty, leading to degradation in discrimination power ({\eg} -49.9\% in PubMedQA). By preserving accuracy and more importantly scoring a higher Cond. AUROC score on all datasets as shown in Table \ref{tab:runner_up_details_vertical}, the RunnerUp strategy forces the proxy task to resolve the specific ambiguity that confuses the model most, effectively acting as a margin-sampling technique \cite{Xie2022-pt}. Additionally, the reduction of variance in the output entropy distribution after statement adaptation makes it a much more stable signal. Thus, we adopt this strategy as the robust choice for general-purpose Active Testing given our choice of $\mathcal{F}_{main} = $ Llama 3.1 8B. We utilize GPT-4o \cite{OpenAI2024-hr} for text reformatting.

\paragraph*{Acquisition Functions}
Our goal is to leverage the uncertainty of the statement-adapted proxy task to perform importance sampling, selecting informative samples earlier in the acquisition process. We process the candidate pool using a main model $\mathcal{F}_{main}$ and, optionally, a surrogate $\mathcal{F}_{surrogate}$. For each sample $x$, we compute a utility score derived from the predictive probability distribution over the fixed binary classes $p(y|x)$. These scores are converted into a Probability Mass Function (PMF) for weighted sampling without replacement.

\begin{table}[h]
\centering
\small
\setlength{\tabcolsep}{2.5pt}
\begin{tabularx}{\linewidth}{@{} l c c c @{}}
\toprule
\textbf{Metric} & \textbf{PubMedQA} & \textbf{MedQA} & \textbf{AI2\_ARC} \\
\midrule
\midrule
Accuracy (QA) & 0.74 & 0.59 & 0.80 \\
Accuracy (Stmt) & 0.77 & 0.64 & 0.74 \\
\midrule
Entropy $\sigma$ (QA) & 0.25 & 0.41 & 0.41 \\
Entropy $\sigma$ (Stmt) & \textbf{0.16} & \textbf{0.12} & \textbf{0.09} \\
\midrule
Cond. AUROC (QA) & 0.693 & 0.687 & 0.769 \\
Cond. AUROC (Stmt) & \textbf{0.707} & \textbf{0.721} & \textbf{0.794} \\

\bottomrule
\end{tabularx}

\caption{\textbf{Impact of RunnerUp Statement Adaptation.} Compared to the QA task, the statement adapted task has a lower spread of entropy $\sigma$ values.}
\label{tab:runner_up_details_vertical}
\end{table}

While standard approaches use raw \textit{Shannon Entropy} or \textit{Upper Confidence Bounds (UCB)} (see Appendix Section \ref{sec:appendix_baselines}), we observe that LLMs often exhibit uncalibrated confidence, leading to being "confidently wrong" on hallucinations. To mitigate this, we propose a family of Regularized Acquisition Functions that interpret uncertainty not as raw entropy, but as divergence from a uniform prior.

\begin{figure*}[htbp]
    \centering
    \begin{minipage}{0.48\textwidth}
        \centering
        \includegraphics[width=\linewidth]{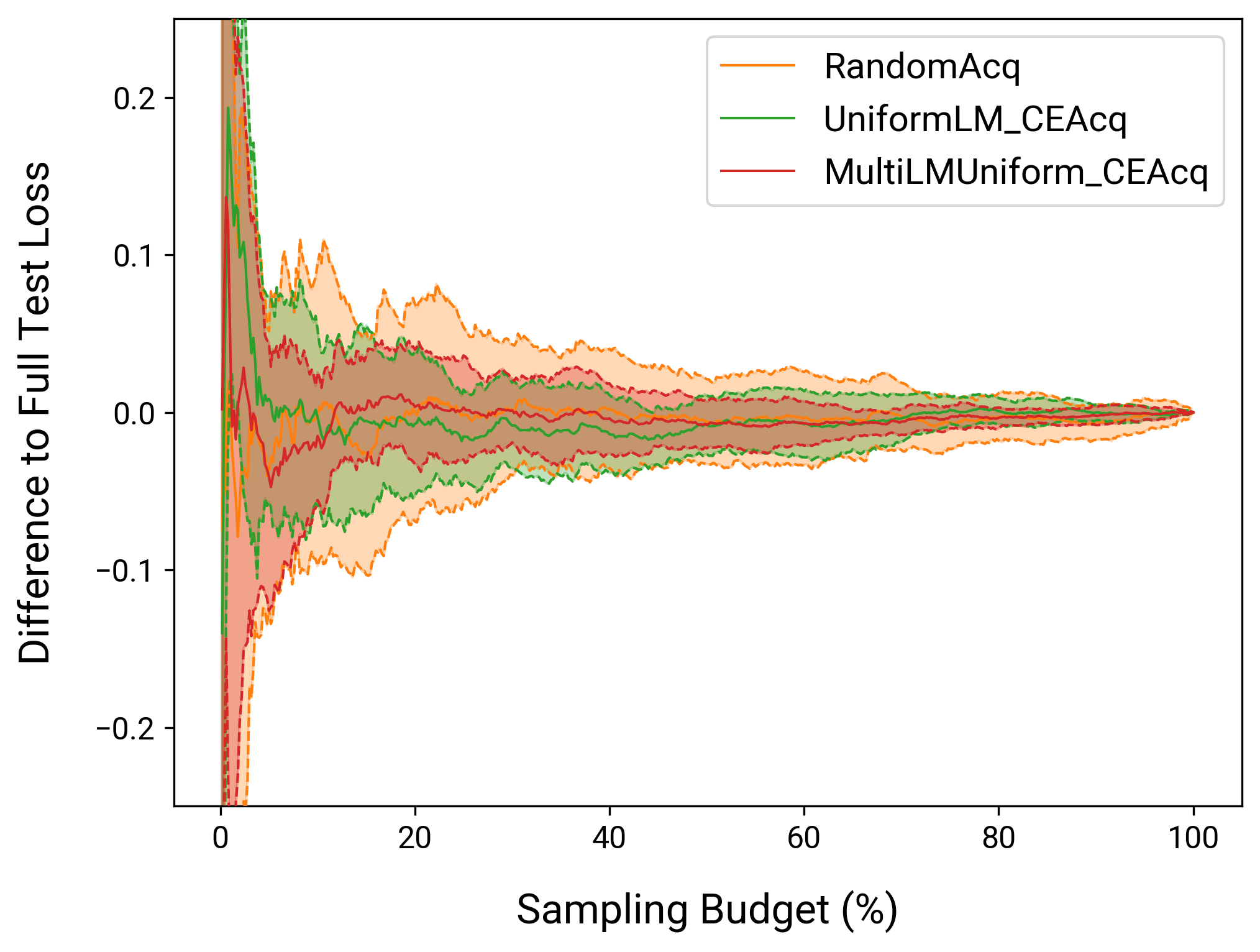}
        {\hypertarget{fig:variance} (a) Variance}
    \end{minipage}
    \hfill
    \begin{minipage}{0.48\textwidth}
        \centering
        \includegraphics[width=\linewidth]{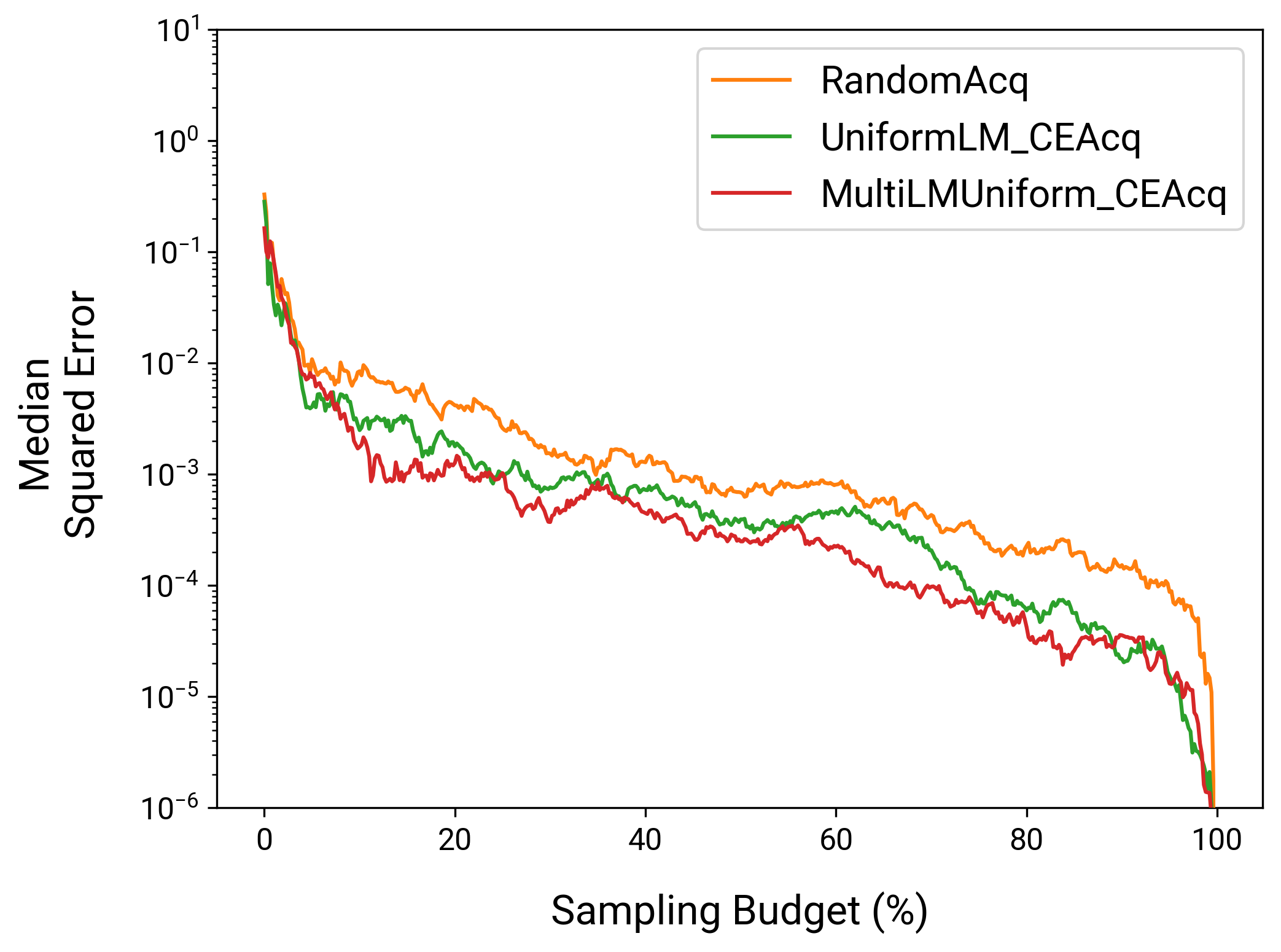}
        {\hypertarget{fig:estimation_error} (b) Average Estimation Error}
    \end{minipage}

    \caption{\textbf{Variance and Average Estimation Error.} Results across acquisition runs on the PubMedQA dataset. Experimental settings: $\mathcal{F}_{\text{main}} = \text{Llama 3.1 8B}$, $\mathcal{F}_{\text{surrogate}} = \text{NLIVerifier}$, and $\mathcal{F}_{\text{oracle}} = \text{RunnerUp}$.}
    
    \label{fig:variance_estimation_error_plots}
    
\end{figure*}

\textit{Zero-Shot Regularization} (\textit{UniformLM\_CEAcq}):
This function measures the cross-entropy between the model's prediction and a uniform distribution $U(y|x) = \frac{1}{K}$. By penalizing deviations from maximum entropy, this effectively down-weights samples where the model is deceptively confident, smoothing the selection distribution.
\begin{equation}
      \mathcal{U}_{Uniform}(x) = -\sum_{i=1}^{K} \frac{1}{K} \log p(y_i|x)
\end{equation}

\textit{Ensemble Regularization} (\textit{MultiLMUniform\_CEAcq}):
We extend this by incorporating a surrogate model ({\eg} an NLV model) to capture distributional disagreement. We calculate the cross-entropy of $\mathcal{F}_{main}$ against a mixture distribution $q^\prime(y|x)$ that combines the surrogate's belief $q(y|x)$ with a uniform prior, balanced by weights $w_1, w_2$:
\begin{equation}
      q^\prime(y_i|x) = w_{1}q(y_i|x) + w_{2}\frac{1}{K}
\end{equation}

Apart from cross-entropy, we also evaluate relative entropy-based surrogate functions utilizing Jensen-Shannon Divergence (JSD) as described in Appendix Section \ref{sec:appendix_baselines}. This hybrid approach captures both epistemic uncertainty (disagreement between models) and aleatoric uncertainty (divergence from uniform), offering a robust signal for active testing.

\section{Experiments}

\paragraph*{Task Settings} 
We perform a structured evaluation of our GAT Framework across different generative tasks: \textit{(RQ2.1) Performance of LLM-based Acquisition on pure classification tasks} and \textit{(RQ2.2) Performance of LLM-based Acquisition on our adapted NLV Task}. Since using models based on the same architecture may not capture the full diversity of the data distribution, we aim to select models with different architectures as surrogates. This way we show whether the uncertainty estimates provided by our proposed acquisition functions work with the default classification-based unbiased risk estimator $\hat{R}_{LURE}$ when adapted to a wide range of generative tasks. In the direct text-classification setting, we evaluate models Llama 3.1 8B and BARTClassifier as both main and surrogate models. With MCQA tasks requiring statement adaptation, we evaluate the performance of the task-equivalent setting: $\mathcal{F}_{oracle} = RunnerUp$ and $\mathcal{F}_{main} =$ Llama 3.1 8B in Table \ref{table:metrics_llama_oracle_runnerup}. 

\begin{table}[h!]
      \centering
      \footnotesize
      \setlength{\tabcolsep}{0.2pt}
      
      \renewcommand{\tabularxcolumn}[1]{m{#1}}
      

          \begin{tabularx}{\linewidth}{@{} l Y Y @{}}
          \toprule
          \textbf{Acquisition Function} & \textbf{AGNews $\mathcal{F}_{main} =$ Llama 3.1 8B} & \textbf{AGNews $\mathcal{F}_{main} =$ BARTClassifier}\\ 
          \midrule
          \midrule
      
          RandomAcq & 0.00095 & 0.00031\\ 

          \midrule
      
          SelfEntropyAcq & 0.00153 & 0.00037\\
      
          UCBAcq & \textbf{0.00044} & 0.00032\\
      
          \midrule
      
          MultiLM\_CEAcq & 0.00096 & \underline{0.00022}\\
      
          UniformLM\_CEAcq & 0.00066 & 0.00024\\
      
          MultiLMUniform\_CEAcq & \underline{0.00045} & 0.00032\\

          \midrule
      
          MultiLM\_JSDAcq & 0.00099 & \textbf{0.00021}\\
      
          UniformLM\_JSDAcq & 0.00096 & 0.00029\\
      
          MultiLMUniform\_JSDAcq & 0.00067 & 0.00028\\

          \midrule
          \midrule

          Performance Gain & \textcolor{MySoftGreen}{+53.68\%} & \textcolor{MySoftGreen}{+32.26\%} \\
          
          \bottomrule
          \end{tabularx}
          \caption{\textbf{Estimation Error AUC ($\downarrow$).} Performance of acquisition functions on traditional text classification datasets such as AGNews.}
          \label{table:metrics_agnews}
\end{table}

\begin{table*}[h!]
    \centering
    \footnotesize
    \renewcommand{\arraystretch}{1} 
    \setlength{\tabcolsep}{3pt} 
        \begin{tabularx}{\linewidth}{@{} l|*3Y : Y Y @{}}
        \toprule
        \textbf{Acquisition Function} & \textbf{PubMedQA} & \textbf{MedQA} & \textbf{AI2\_ARC} & \textbf{Geo Mean AUC} \\ 
        \midrule
        \midrule
    
        RandomAcq & 0.00339 & 0.00061 & 0.00058 & 0.00106 \\ 

        \midrule
    
        SelfEntropyAcq & 0.01526 & 0.00055 & 0.00094 & 0.00199 \\
    
        LMClusterEntropyAcq & 0.00531 & 0.00117 & \underline{0.00049} & 0.00145 \\
    
        UCBAcq & 0.01772 & 0.00099 & 0.00069 & 0.00230 \\
    
        \midrule
    
        MultiLM\_CEAcq & 0.00425 & 0.00051 & 0.00126 & 0.00140 \\
    
        UniformLM\_CEAcq & \underline{0.00160} & \textbf{0.00041} & \textbf{0.00039} & \textbf{0.00064} \\
    
        MultiLMUniform\_CEAcq & \textbf{0.00112} & \underline{0.00048} & 0.00060 & \underline{0.00069} \\

        \midrule
    
        MultiLM\_JSDAcq & 0.11365 & 0.15805 & 0.00710 & 0.05034 \\
    
        UniformLM\_JSDAcq & 0.03877 & 0.01035 & 0.01716 & 0.01903 \\
    
        MultiLMUniform\_JSDAcq & 0.00574 & 0.00457 & 0.00277 & 0.00417 \\
        
        \midrule
        \midrule
        
        Performance Gain & \textcolor{MySoftGreen}{+66.96\%} & \textcolor{MySoftGreen}{+32.79\%} & \textcolor{MySoftGreen}{+32.76\%} & \textcolor{MySoftGreen}{+39.62\%} \\
        
        \bottomrule
        \end{tabularx}
        \caption{\textbf{Estimation Error AUC ($\downarrow$).} Performance of acquisition functions, where the geometric mean highlights robustness across multiple datasets. Experimental settings: $\mathcal{F}_{\text{main}} = \text{Llama 3.1 8B}$, $\mathcal{F}_{\text{surrogate}} = \text{NLIVerifier}$, and $\mathcal{F}_{\text{oracle}} = \text{RunnerUp}$.}
        \label{table:metrics_llama_oracle_runnerup}
    \end{table*}

\paragraph*{Datasets}
For (RQ2.1), we employ the dataset AGNews \cite{Zhang2015-xh} to establish a baseline for our acquisition functions in a traditional classification setting. In GAT (RQ2.2), we prioritize tasks that demand \textit{context-based reasoning} over simple factual recall. We select three Multiple-Choice QA datasets: PubMedQA \cite{Jin2019-au} (Extractive Reasoning), MedQA \cite{Jin2020-zk} (Clinical Reasoning), and AI2\_ARC \cite{Clark2018-tl} (Scientific Reasoning). These benchmarks were chosen because they require the model to synthesize information from specific contexts, ranging from clinical patient vignettes to complex scientific scenarios, thereby rigorously testing the model's inferential stability under uncertainty. Detailed statistics for each dataset are provided in Appendix Section \ref{sec:dataset-statistics}.

\paragraph*{Metrics}
To evaluate the efficiency of our Active Testing framework (RQ2), we assess how quickly the estimated model performance converges to the true performance as the annotation budget increases. We define the Estimation Error as the Squared Error (SE) between the estimated risk ($\hat{R}_S$) derived from the selected subset and the true risk ($R_{true}$) calculated on the full test set. We compute this error across a range of sampling budgets $b \in [5\%, 50\%]$ and integrate the resulting curve to produce the Estimation Error Area Under the Curve (AUC) \cite{Huang2026-we}. A lower Estimation Error AUC indicates that the acquisition function identifies a representative subset of samples more rapidly than random sampling, directly quantifying the reduction in expert annotation effort required to achieve reliable benchmarking.

\section{Findings and Discussion}

\paragraph*{Surrogate-based Acquisition in Text Classification}
Looking at the performance of acquisition functions in Table \ref{table:metrics_agnews} on the AGNews dataset, we see that \textit{UniformLM\_CEAcq} and \textit{MultiLM\_JSDAcq} outperforms our baseline \textit{RandomAcq} for both variations of $\mathcal{F}_{main}$ and $\mathcal{F}_{surrogate}$. We also observe that including a uniform prior is much more effective for $\mathcal{F}_{main} =$ Llama 3.1 8B (53.7\%) compared to $\mathcal{F}_{main} =$ BARTClassifier (22.5\%). This shows that regularization mitigates mis-calibration of entropy distributions for non-discriminative models such as LLMs while models finetuned for discriminative tasks have more stable decision boundaries. These results support our hypothesis on the extendibility of LLMs for Active Testing (RQ2.1).

\paragraph*{Acquisition Performance on Proxy Classification Task}
Leveraging the stabilized uncertainty profile of the \textit{RunnerUp} proxy task, our proposed acquisition functions demonstrate substantial efficiency gains over traditional baselines. As detailed in Table \ref{table:metrics_llama_oracle_runnerup}, \textit{UniformLM\_CEAcq} emerges as the robust winner, achieving a {39.6\%} reduction in geometric mean estimation error compared to \textit{RandomAcq}. Similarly, \textit{MultiLMUniform\_CEAcq} yields a 34.9\% reduction, validating its efficacy across diverse datasets. Crucially, the comparison reveals the necessity of our regularization approach: \textit{SelfEntropyAcq} and \textit{UCBAcq} perform significantly \textit{worse} than random sampling. This failure of uncalibrated methods shows that raw LLM confidence is often unreliable as visualized in Figure \ref{fig:variance_estimation_error_plots}. 

\paragraph*{Cross-Entropy vs Entropy Divergence}
In our results, we observe that the best-performing acquisition functions in our proxy task utilize Cross-Entropy over Entropy Divergence-based metrics to predict uncertainty. This indicates that cross-entropy more closely signifies the true expected loss for LLMs and in a given test set. When looking into each $\mathcal{F}_{main}$ and $\mathcal{F}_{surrogate}$ pairs, we observe that when using a classification-based model as $\mathcal{F}_{main}$ such as NLIVerifier or BARTClassifer, MultiLMUniform\_JSDAcq outperforms the baseline on all oracle strategies. This shows the impact that task-specific customization and logit calibration can have on our choice of loss/distance function selection for LLM-based acquisition.

\section{Conclusion}
In this work, we introduce \textbf{Generative Active Testing (GAT)}, a framework that addresses the bottleneck of expert annotation in high-stakes domains by converting dynamic MCQ tasks into fixed \textit{Statement Verification} proxies. We demonstrate that our novel \textit{RunnerUp} adaptation strategy effectively stabilizes uncertainty estimates and targets the model's decision boundaries, significantly improving error detection in the critical high-confidence samples. By leveraging this proxy task with \textit{zero-shot, uniform-prior acquisition functions} ({\eg} \textit{UniformLM\_CEAcq}), we achieve around \textbf{40\%} reduction in estimation error AUC compared to random sampling. These results offer a scalable pathway for reliable model benchmarking in industrial healthcare applications, substantially reducing the volume of data requiring expert clinician review without the need for expensive fine-tuning.

\section*{Limitations}
In our current setup for GAT, we only utilize zero-shot surrogates for providing sample-level uncertainty estimates. Finetuning surrogates on acquired samples to inform the selection of future candidates is a future direction we wish to pursue. Secondly, although our evaluation includes datasets that handle different types of QA tasks, the answer format is limited to choosing an option from multiple-choice candidates. We aim to explore how our adaptation module can be extended for Active Testing on open-domain QA and other long-form generation tasks.

\bibliography{custom}

\clearpage
\appendix

\section*{Appendix}
\label{sec:appendix}

\section{Ablation Studies}
\label{sec:ablation-studies}

\paragraph*{Evaluation of Clustering-based Acquisition Functions}
To evaluate the efficacy of structural diversity in sampling, we also evaluate a hierarchical clustering approach inspired by \cite{Huang2026-we}. We first generate embeddings for the entire test set using Llama 3.1 8B and group them using balanced k-means clustering to ensure non-sparse partitions. The acquisition function operates in two stages to reduce computational overhead. In each step, we first identify the set of valid clusters (those containing remaining samples) and select one uniformly at random, $c \sim U(C_{\text{valid}})$. We then compute the predictive entropy only for the samples within the selected cluster $c$ to form a local probability mass function $P(x|c)$. The final sample is drawn from this local distribution, and the global inclusion probability is calculated as $\pi_t = P(x|c) \cdot P(c)$.

Table \ref{table:metrics_cluster_entropy_llama} highlights a critical bias-variance trade-off in this hierarchical setting. The Unbiased Estimator, which utilizes the calculated weights $\pi_t$ for risk estimation (LURE), exhibits high estimation error ({\eg} 0.00531 on PubMedQA). This is attributed to the "variance explosion" inherent in two-stage sampling: the compounding of cluster selection probability ($1/|C|$) and local sampling probability creates extremely small $\pi_t$ values, resulting in volatile importance weights. In contrast, the Biased Estimator (Simple Mean), which ignores these weights, achieves significantly lower error (0.00166) by sacrificing theoretical unbiasedness for variance reduction.

\begin{table*}[h!]
    \centering
    \footnotesize
    \renewcommand{\arraystretch}{1} 
    \setlength{\tabcolsep}{3pt}    
    \begin{tabularx}{\linewidth}{@{} l|*3Y @{}}
    \toprule
    \textbf{Acquisition Function} & \textbf{PubMedQA} & \textbf{MedQA} & \textbf{AI2\_ARC} \\ 
    \midrule
    \midrule
  
    RandomAcquisition & 0.00339 & \textbf{0.00061} & 0.00058 \\ 
  
  
    \midrule
    \midrule
  
    LMClusterAcquisition (Unbiased Estimator) & 0.00531 & 0.00117 & \textbf{0.00049} \\ 
  
    LMClusterAcquisition (Biased Estimator) & \textbf{0.00166} & 0.00084 & 0.00062 \\      
  
    \bottomrule
    \end{tabularx}
    \caption{\textbf{Estimation Error AUC ($\downarrow$).} Performance of different Cluster Acquisition functions. Experimental settings: $\mathcal{F}_{\text{main}} = \text{Llama 3.1 8B}$, $\mathcal{F}_{\text{surrogate}} = \text{NLIVerifier}$, and $\mathcal{F}_{\text{oracle}} = \text{LeastConf}$.}
    \label{table:metrics_cluster_entropy_llama}
\end{table*}

\paragraph*{Importance of Statement Adaptation}
With our unbiased estimators functioning effectively for classification-style tasks, we analyzed whether they would be effective directly for MCQA sample selection without Statement Adaptation. In PubMedQA, all questions have the same options, allowing us to test this hypothesis and validate whether Statement Adaptation is crucial for informed sample selection. Here, we consider every option index to be a unique class and model the entropy to measure the uncertainty of selecting a particular index as the answer. From our results in Table \ref{table:metrics_noadaptation}, we observe that Active Testing performance is significantly worse compared to simple Random Sampling across all acquisition functions tested, even when the options for each question are seemingly the same. Our ablation experiments underscore the effectiveness of the Statement Adaptation module as a key component of the GAT pipeline, enabling the accurate quantification of uncertainty across QA tasks with varying question and option types.

\begin{table}[t]
\centering
\small
\setlength{\tabcolsep}{2.5pt}
\begin{tabularx}{\linewidth}{@{} l c c c @{}}
\toprule
\textbf{Metric} & \textbf{PubMedQA} & \textbf{MedQA} & \textbf{AI2\_ARC} \\
\midrule
\midrule
Accuracy (QA) & 0.75 & 0.59 & 0.80 \\
Accuracy (Stmt) & 0.44 & 0.77 & 0.82 \\
\midrule
Entropy $\sigma$ (QA) & 0.25 & 0.41 & 0.41 \\
Entropy $\sigma$ (Stmt) & \textbf{0.18} & \textbf{0.18} & \textbf{0.12} \\
\midrule
Cond. AUROC (QA) & \textbf{0.693} & 0.687 & 0.769 \\
Cond. AUROC (Stmt) & 0.347 & \textbf{0.863} & \textbf{0.853} \\
\bottomrule
\end{tabularx}
\caption{\textbf{Impact of Least Confident Statement Adaptation.} Unlike the RunnerUp strategy, picking the least confident option results in high variance. While it improves discrimination on MedQA and AI2\_ARC, it causes a catastrophic collapse in performance on PubMedQA (AUROC 0.69 $\rightarrow$ 0.35), making it unreliable for general deployment.}
\label{tab:least_conf_details}
\end{table}

\begin{table}[t]
\centering
\small
\setlength{\tabcolsep}{2.5pt}
\begin{tabularx}{\linewidth}{@{} l c c c @{}}
\toprule
\textbf{Metric} & \textbf{PubMedQA} & \textbf{MedQA} & \textbf{AI2\_ARC} \\
\midrule
\midrule
Accuracy (QA) & 0.75 & 0.59 & 0.80 \\
Accuracy (Stmt) & 0.69 & 0.60 & 0.43 \\
\midrule
Entropy $\sigma$ (QA) & 0.25 & 0.41 & 0.41 \\
Entropy $\sigma$ (Stmt) & \textbf{0.23} & \textbf{0.10} & \textbf{0.07} \\
\midrule
Cond. AUROC (QA) & 0.693 & \textbf{0.687} & \textbf{0.769} \\
Cond. AUROC (Stmt) & \textbf{0.720} & 0.645 & 0.390 \\
\bottomrule
\end{tabularx}
\caption{\textbf{Impact of Most Confident Statement Adaptation.} Verifying the model's top choice fails to consistently improve calibration. On AI2\_ARC, the method suffers a severe collapse (AUROC 0.77 $\rightarrow$ 0.39), and accuracy drops significantly, confirming that this strategy is prone to confirming the model's original errors.}
\label{tab:most_conf_details}
\end{table}

\begin{table}[!h]
    \centering
    \scriptsize
        \begin{tabularx}{\linewidth}{@{} c|Y @{}}
        \toprule
        Model & Model ID \\
        \midrule
        \midrule

        Llama 3.1 8B & \texttt{meta-llama/Meta-Llama-3-8B} \\
        NLI Verifier & \texttt{soumyasanyal/nli-entailment-verifier-xxl} \\
        GPT-4o & \texttt{gpt-4o-2024-07-01-preview (Azure OpenAI)} \\
        
        \bottomrule
        \end{tabularx}
       \caption{\textbf{Overview of models evaluated for Active Testing.} For open-source models, this table shows the model names in Huggingface.}
        \label{table:llm_ids}
\end{table}

\section{Differentiation of Active Testing from Active Learning}
\label{sec:active-learning-testing-diff}

Active Testing is a well defined topic in the field of sample-efficient evaluation, as evidenced by past research with predictive Machine Learning models \cite{Kossen2021-xc, Sawade2010-cr, Ochiai2023-ib}. Our work primarily attempts to extend this setting with Language Models (LM) when the tasks are generative in nature. Active testing is fundamentally different from active learning because it focuses on accurately estimating a model’s performance, not improving it. Unlike active learning, which avoids noisy (aleatoric) regions and can tolerate biased estimates, active testing often needs to sample from those uncertain areas and requires unbiased, low-variance error estimation \cite{Kossen2021-xc}. This is our main motivation behind adapting theoretically sound unbiased estimators for generative tasks in GAT.

\section{Dataset Statistics}
\label{sec:dataset-statistics}

\paragraph*{PubMedQA}
PubMedQA is a biomedical question-answering dataset derived from PubMed abstracts, designed to answer research questions with "yes," "no," or "maybe" responses. The dataset contains 1,000 expert-annotated QA instances, 61,200 unannotated instances, and 211,300 artificially generated QA instances. Similar to other approaches, we use 500 samples from the 100 expert-annotated samples for active sample selection. The uniqueness of PubMedQA lies in its deep mining of PubMed data sources. Approximately 760,000 PubMed article titles are presented in the form of questions, which are often directly related to the conclusion part of the abstract, providing a direct answer source for QA systems. 

\paragraph*{MedQA}
MedQA is a multiple-choice medical question-answering dataset sourced from medical licensing exams in the United States, mainland China, and Taiwan, which evaluate physicians' professional knowledge and clinical decision-making skills. Answering questions about a wide range of medical topics requires a deep understanding of relevant concepts. The dataset includes 61,097 questions, with 12,723 in English, 34,251 in Simplified Chinese, and 14,123 in Traditional Chinese. The dataset is available in both its original form and a refined version with four answer choices. For our experiments, we use the test set with 1273 examples for active sample selection.

\paragraph*{AI2 Reasoning Challenge (ARC)}
ARC is a multiple-choice QA dataset, containing 7,787 genuine grade-school level science questions. The dataset has two partitions, notably the Easy and Challenge set, where the latter has questions that were incorrectly answered by both an retrieval-based and word co-occurrence algorithm. For our experiments, we exclusively focus on the challenge set, and filter the test set based off whether the question has multiple sentences or not. This is done to ensure that we segregate the context away from the actual question, consistent with our evaluation strategy and prompt templates for other datasets. By this filtering, we remove $\sim$52\% of samples and perform Active Testing on 616 samples for our experiments.

\paragraph*{AGNews}
AGNews is a text classification dataset comprising a total of 127,600 news articles gathered from more than 2000 news sources. All the articles were collected using ComeToMyHead, which is an academic news search engine that has been running since July 2004. The dataset includes 120,000 training samples and 7,600 test samples, each categorized into one of four distinct classes: World, Sports, Business, and Science/Technology. Each entry in the dataset consists of a short news article accompanied by its title. The dataset is provided by the academic community for research purposes in data mining (clustering, classification, etc), information retrieval (ranking, search, etc), XML, data compression, data streaming, and any other non-commercial activity. We perform Active Testing on the test set containing 7600 samples for our experiments.

\begin{table}[h!]
\centering
    \footnotesize
    \begin{tabularx}{\linewidth}{@{} l|Y @{}}
    \toprule
    \textbf{Acquisition Function} & \textbf{PubMedQA}\\ 
    \midrule
    \midrule

    RandomAcq & \textbf{0.01289} \\ 

    AccuracyAcq & 0.03950 \\
    SelfEntropyAcq & 0.06692 \\

    UCBAcq & \underline{0.02147} \\

    \midrule
    \midrule

    UniformLM\_CrossEntropyAcq & 0.02337 \\

    UniformLM\_JSDAcq & 0.03831 \\
    
    \bottomrule
    \end{tabularx}
    \caption{\textbf{Estimation Error AUC ($\downarrow$).} Performance of acquisition functions without statement adaptation on the PubMedQA dataset. Experimental settings: $\mathcal{M}_{\text{main}} = \text{Llama 3.1 8B}$.}
    \label{table:metrics_noadaptation}
\end{table}

\section{Baseline Acquisition Functions}
\label{sec:appendix_baselines}

In addition to our proposed regularized functions, we evaluate several standard uncertainty estimation baselines and divergence metrics.

\paragraph*{SelfEntropyAcq \& UCBAcq}
The most direct measure of uncertainty is the raw \textbf{Shannon Entropy} of the predictive distribution:
\begin{equation}
    H(y|x) = -\sum_{i=1}^{m} p(y_i|x) \log p(y_i|x)
\end{equation}
To balance exploration and exploitation, we also employ an \textbf{Upper Confidence Bound (UCB)} strategy:
\begin{equation}
      UCB(y|x) = \mu(y|x) + \beta \cdot \sigma(y|x)
\end{equation}
where $\beta = \sqrt{\log(2/n^{-9/2})}$ is a hyperparameter dependent on dataset size $n$.

\paragraph*{Surrogate-Based Divergence}
To capture epistemic uncertainty without uniform regularization, we employ \textit{MultiLM\_CEAcq}, which computes the cross-entropy between the main model $p$ and a surrogate model $q$:
\begin{equation}
      H_{ML}(y|x) = -\sum_{i=1}^{m} q(y_i|x) \log p(y_i|x)
\end{equation}

\paragraph*{Jensen-Shannon Divergence (JSD)}
Finally, we evaluate symmetric divergence measures. \textit{MultiLM\_JSDAcq} calculates the JSD between the main and surrogate distributions, capturing uncertainty as bidirectional disagreement:
\begin{equation}
      JSD(P || Q) = \frac{1}{2} \left( D_{KL}(P || M) + D_{KL}(Q || M) \right)
\end{equation}
where $M$ is the average distribution $\frac{P+Q}{2}$. We also extend this to \textit{MultiLMUniform\_JSDAcq} (Eq. \ref{eq:jsd_uniform}) which calculates divergence between the main, surrogate, and uniform distributions:
\begin{multline}
\label{eq:jsd_uniform}
    JSD(P || Q || U) = \frac{1}{3} [ D_{KL}(P || M^\prime) \\ 
    + D_{KL}(Q || M^\prime) + D_{KL}(U || M^\prime) ]
\end{multline}

\section{Prompt Templates and Evaluation Settings}
\label{appendix:prompt_templates}

For all our acquisition functions, we report the average of 10 runs on all datasets compared. We use a custom random generator with an initial seed set to 42 for all our experiments. We use a template QA prompt for Llama 3.1 8B on both Classification and NLV tasks as described below.
\begin{figure*}[!ht]
    \centering
    \footnotesize
    \begin{tcolorbox}[title={Llama 3.1 8B (Default QA)}, colframe = orange!30, colback = orange!10, coltitle = orange!20!black, after skip=0pt, boxsep=5pt, width=\linewidth]

    Context: \red{<insert-context>} \\
    Question: \red{<insert-question>} \\
    Choices: \red{\{ A. <insert-option1>, B. <insert-option2>, ... \}} \\
    Answer: 
    
    \end{tcolorbox}
\end{figure*}

\begin{figure*}[!ht]
    \centering
    \footnotesize
    \begin{tcolorbox}[title={Llama 3.1 8B (Natural Language Verification)}, colframe = orange!30, colback = orange!10, after skip=0pt, coltitle = orange!20!black, boxsep=5pt, width=\linewidth]

    Context: \red{<insert-context>} \\
    Statement: \red{<insert-statement>} \\
    \red{Is the claim supported by the given context? Return only the option corresponding to the correct choice.} \\
    Choices: \red{\{ A. true, B. false \}} \\
    Answer: 
    
    \end{tcolorbox}
\end{figure*}

\begin{figure*}[!ht]
    \centering
    \footnotesize
    \begin{tcolorbox}[title={Llama 3.1 8B (AGNews)}, colframe = orange!30, colback = orange!10, coltitle = orange!20!black, after skip=0pt, boxsep=5pt, width=\textwidth]

    User: \red{Please help me perform a news classification task. I will give you a news title and the corresponding description. You should classify the news into the categories of "World", "Sports", "Business", and "Technology". You are only allowed to give me a word, selecting from these four categories.} \\
    Context: \red{<insert-context>} \\
    Category: \red{ \{ A. World, B. Sports, C. Business, D. Technology \} } \\
    Answer:
    
    \end{tcolorbox}
\end{figure*}

\begin{figure*}[!ht]
    \centering
    \footnotesize
    \begin{tcolorbox}[title={NLIVerifier (Natural Language Verification)}, colframe = green!30, colback = green!10, coltitle = green!20!black, after skip=0pt, boxsep=5pt, width=\textwidth]

    Premise: \red{<insert-premise>} \\
    Hypothesis: \red{<insert-hypothesis>} \\
    \red{Given the premise, is the hypothesis correct?} \\
    Answer:
    
    \end{tcolorbox}
\end{figure*}

\end{document}